# $\mathcal{RES}$—A Relative Method for Evidential Reasoning

Zhi An, David A. Bell, John G. Hughes
Department of Information Systems
University of Ulster at Jordanstown
Newtownabbey, Co. Antrim
BT37 0QB, N. Ireland, UK.
*E-mail:CBFC23@UJVAX.ULSTER.AC.UK*

## Abstract

In this paper we describe a novel method for evidential reasoning [1]. It involves modelling the process of evidential reasoning in three steps, namely, evidence structure construction, evidence accumulation, and decision making.

The proposed method, called $\mathcal{RES}$, is novel in that evidence strength is associated with an evidential support relationship (an argument) between a pair of statements and such strength is carried by comparison between arguments. This is in contrast to the onventional approaches, where evidence strength is represented numerically and is associated with a statement.

## 1 Introduction

> ...*talking about 'good' measurements and 'bad' ones...*
> *does not any* **analysis** *of measurement require concepts more* **fundamental** *than measurement? And should not the fundamental theory be about these more fundamental concepts?*
>
> —*J.S. Bell, 'Quantum Mechanics for Cosmologists'*

It is sometimes difficult enough to make a judgement even if we can observe directly that which we want to judge. But there are many uncertain situations in which we cannot make observations directly. Instead, in such situations, we have to resort to arguments. Arguments, while not directly available from observations, link observable facts to unobservable features. And these arguments are then taken as our justifications for judgements. Obviously, in our reasoning process, good arguments give excellent justification for our statements, second only to observable facts. Thus, arguments should be a focus for studies of uncertain reasoning.

But in uncertain situations, it is also commonly found that many arguments are presented and that these arguments might be inconsistent. They might be based on different observable facts or derived from different parts of our knowledge or oriented towards different points of view. In such situations, we have to address the strengths of these arguments for competitive statements to resolve any inconsistency. A common activity in such situations is to compare the arguments supporting and refuting competitive statements respectively. Usually, arguments for competitive statements are weighed with respect to each other, *i.e.* the arguments are compared with each other with regard to how trustworthy they are. Then, we make our judgements according to the results of these comparisons.

It is very desirable to have some standard for measurement so that we can measure the strengths of all these arguments to find out which statement we should choose as our decision, based on many arguments. But because there are many factors affecting the strengths of arguments, such measures are not easily available and are sometimes not suitable. Moreover, such measures may not be necessary in that sometimes only comparisons among arguments are sufficient.

In this paper we describe a method for evidential reasoning based on reflecting the above ideas. The method is called $\mathcal{RES}$ for Relative Evidential Support. In $\mathcal{RES}$, evidential support relationships between statements, called *arguments*, and their relative strengths, represented by comparisons between arguments, are represented.

In the following section, we present our viewpoint that evidential reasoning is a process composed of three steps, namely, evidence structure construction, evidence accumulation, and decision making. In section 2.1, the steps are modelled in the method $\mathcal{RES}$. A simple example is presented along with the exposition of the method. In section 3, a well known example in the literature is analyzed using $\mathcal{RES}$, which provides a basis for us to compare $\mathcal{RES}$ against probabilistic methods and the endorsement model. In the last section, a brief summary is given. Some other results of our studies of $\mathcal{RES}$ are also listed without going into details.

## 2  Our Point of View

For pragmatic purposes, human reasoners must reject the assertion that *nothing is true but The Whole*[1], where *The Whole* is the complete truth of the world. We understand *something* and we have *some* knowledge. But at the same time, there is much about the world and about The Whole that we do not know or we do not know exactly. We have to make judgements based on that which we do know, and act on these judgements. The situation is like that in which we are observing a distant object and want to figure out its features. Because it is distant, some of its features cannot be observed directly. At the outset, we must know and understand some of the relationships between our observations and the features of the object. Otherwise our observations are useless. Sometimes, we might not really *know* a relationship, but we *believe* it. At the same time, we know that some of the relationships are very accurate (or convincing) and some of them are less accurate. Generally speaking, these relationships have strengths of some kind.

With knowledge as described above, we can make our observations. We are guided at this time by our knowledge that we are looking for observations which *we know or believe* can be related to the features of the object. Other observations should not be made, because they are simply useless for the task in hand.

As the observations are made, some of our knowledge may be triggered to give us justifications for predicating the features of the object. During this process, some predications might conflict with others. In such cases, we will have to compare the justifications supporting these conflicting predications. As a result, some predications with weak justifications might be overwhelmed by others with stronger justifications.

We view evidential reasoning as a process as described above. Concretely, we view evidential reasoning to be a process composed of three steps, namely, evidence structure construction, evidence accumulation, and decision making. In the first step, we collect what we know about the relationships between that which is observable and that which we want to figure out. Judging the strengths of these relationships is a very important part of this step. In the second step we make our observations and fit them into the structure. In the last step, we make our decisions based on the result of the first two steps.

In the following section, we propose a model of evidential reasoning which reflects these steps directly.

### 2.1  $\mathcal{RS}$–A Model for Relative Evidential Support

The semantic model for the method is based on two abstract spaces for making judgements (statements). One space is for sensing evidence and the other is for making decisions.[2] Here we formalize the two spaces as two first order logics. We denote the space for evidence sensing as $\mathcal{L}_e$ and the space for decision making as $\mathcal{L}_p$.[3]

Evidence (consisting of sentences in the first space) is related to choices (sentences in the second space) only via *arguments* such as "this evidence sentence supports (or refutes) that conclusion sentence". We denote an argument that "evidence $e$ supports choice $p$" as $\langle e, p \rangle$ and will call such pairs "arguments from $\mathcal{L}_e$ to $\mathcal{L}_p$".[4] In $\langle e, p \rangle$, we will call $e$ the *presumption* and $p$ the *conclusion*. Thus, the word "argument" is taken to have the intuitive meaning "a step in reasoning" rather than a special kind of statement.

It should be noticed that all arguments are conditioned in that an argument will not be enforced unless its presumption is satisfied. Thus, an argument $\langle e, p \rangle$ should be read as "if the presumption $e$ is satisfied, then there is an argument supporting the conclusion $p$."

### 2.2  The Structure for Evidence

*Definition:* Let $\mathcal{A}$ be a set of arguments from $\mathcal{L}_e$ to $\mathcal{L}_p$ and $\mathcal{R}$ be a relation on $\mathcal{A}$, denoted by "$\preceq$". Then, the 4-tuple $\mathcal{ES} = \langle \mathcal{L}_e, \mathcal{L}_p, \mathcal{A}, \mathcal{R} \rangle$ is called an *evidence structure* if and only if

1. $\forall \langle e, p \rangle \in \mathcal{A}$,
$$\langle e, p \rangle \preceq \langle e, p \rangle;$$

2. $\forall \langle e_1, p_1 \rangle, \langle e_2, p_2 \rangle, \langle e_3, p_3 \rangle \in \mathcal{A}$, if $\langle e_1, p_1 \rangle \preceq \langle e_2, p_2 \rangle$ and $\langle e_2, p_2 \rangle \preceq \langle e_3, p_3 \rangle$, then
$$\langle e_1, p_1 \rangle \preceq \langle e_3, p_3 \rangle;$$

3. $\forall \langle e, p_1 \rangle, \langle e, p_2 \rangle \in \mathcal{A}$, if $p_1 \rightarrow p_2$ ($p_1$ implies $p_2$ in $\mathcal{L}_p$), then [5]
$$\langle e, p_1 \rangle \preceq \langle e, p_2 \rangle;$$

4. $\forall \langle e_1, p_1 \rangle, \langle e_2, p_2 \rangle \in \mathcal{A}$, if $(e_1 \rightarrow e_2) \wedge \neg (e_2 \rightarrow e_1)$ ($e_1$ implies $e_2$ and $e_1$ is not equivalent to $e_2$ in $\mathcal{L}_e$), then
$$\langle e_2, p_2 \rangle \preceq \langle e_1, p_1 \rangle.$$

---

[1] Though the philosophy of this assertion has its justification. See Russell [7] for his description of Hegel's dialectic.

[2] This division is unnecessary. We retain it because it helps clarify the process of evidential reasoning.

[3] We suggest that $\mathcal{L}_e$ and $\mathcal{L}_p$ be delineated in such a manner that the truth of sentences in $\mathcal{L}_e$ are readily available and the truth of sentences in $\mathcal{L}_p$ are what we are seeking.

[4] We omit arguments such as "evidence $e$ refutes choice $p$" because they can be represented using $\langle e, p \rangle$ in a natural way, as we will show later.

[5] Notice that the two arguments have a common presumption.

5. $\forall \langle e_1, p_1 \rangle, \langle e_2, p_2 \rangle \in \mathcal{A},$

$$\langle e_1 \vee e_2, p_1 \vee p_2 \rangle.$$

It should be noticed that the constraints on evidence structure are very simple. These constraints guarantee that our structure of evidence is consistent. We believe this requirement is necessary so that even if our observations and their implications should be inconsistent in themselves or inconsistent with our knowledge, our knowledge itself should be consistent.

These constraints have intuitive appeal. The first two constraints say that "$\preceq$" is a partial order relation on $\mathcal{A}$. The third constraint requires that under the same presumption, a statement always commands no more support than any of its logical implications. If statements are viewed as subsets of the set of possible values for a variable, this constraint simply states that, by the same evidence, no subset can be supported better than its supersets.

The fourth constraint specifies that any logical implicaiton of a statement cannot enforce stronger support than the statement itself. This constraint is included because when $(e_1 \rightarrow e_2) \wedge \neg(e_2 \rightarrow e_1)$, we can say that $e_1$ is more specific than $e_2$ and contains more information. In such cases, we can say that information conveyed in $e_2$ is contained in that conveyed by $e_1$. Thus, all arguments based on $e_1$ can be viewed as based on something which has already taken into account everything $e_2$ has to say. In this sense, the evidence space can be viewed as being *layered*—any thing said at a higher level contains or covers all things which has been said at a lower level. $(e_1 \rightarrow e_2) \wedge \neg(e_2 \rightarrow e_1)$ means that $e_1$ is at a higher level than $e_2$. Thus, what said by $e_2$ should be weaker than that said by $e_1$ in the sense that $e_1$ might have nothing more to say than $e_2$, but if it does, these things are more defensible than those based on $e_2$. This includes two kinds of cases. In cases where these additional things based on $e_1$ are consistent with those based on $e_2$, knowing $e_1$ enforces what has been said. In cases where the additional things are inconsistent with those bases on $e_2$, the ones based $e_1$ will overrule those based on $e_2$ because more considerations are commanded based on $e_1$.

For example, penguins do not fly though birds do. Because when we are talking about penguins, their being birds has already been considered, then any conclusions reached from penguin considerations have priority over these from *only* bird considerations. In this case, inconsistency arises but the desirable result should be that penguins don't fly based on the more specific consideration of penguins.

For another example, penguins have legs, so do birds. This is a consistent situation where penguin's having legs is supported by their being penguins and being birds. But obviously, no changes concerning only the legs of birds in general can affect this assertion. In this sense, because more specific considerations are available, less specific considerations are overshadowed and become irrelevant.

Notice the symmetry in the last two conditions that *the stronger the evidence* but *the weaker the conclusion, the stronger the argument*.

The last constraint concerns only the *presence* of arguments, *i.e.* the existence of evidential supports relationships. The constraint says that for two arguments, the disjunction of their presumptions supports the disjunction of their conclusions. In other words, it says that if one of the two presumptions is granted, then one of the two conclusions is supported. Notice that, by constraint 4, the argument with the disjunctions is less trustworthy than the original two arguments.

In the relation $\mathcal{R}$ of an evidence structure $\mathcal{ES}$, two arguments might not be related. But if they are, $\langle e_1, p_1 \rangle \preceq \langle e_2, p_2 \rangle$ should be read as "argument $\langle e_1, p_1 \rangle$ is no more believable or no stronger than argument $\langle e_2, p_2 \rangle$".

Some conventions can be used to denote other relationships between arguments. For example, in the following, we will use $\langle e_1, p_1 \rangle \simeq \langle e_2, p_2 \rangle$ for $(\langle e_1, p_1 \rangle \preceq \langle e_2, p_2 \rangle) \wedge (\langle e_2, p_2 \rangle \preceq \langle e_1, p_1 \rangle)$ and $\langle e_1, p_1 \rangle \prec \langle e_2, p_2 \rangle$ for $(\langle e_1, p_1 \rangle \preceq \langle e_2, p_2 \rangle) \wedge \neg(\langle e_2, p_2 \rangle \preceq \langle e_1, p_1 \rangle)$.

**Example 1** *Let $\{Al_1, Al_2, Al_3\}$ be all the alternative states of a system. Suppose it cannot be directly detected what state the system is in. But we can conduct some tests on the system:*
*Test1: $e_1$, if positive, supports choosing alternative $Al_1$ as well as $Al_2$.*
*Test2: $e_2$, if positive, refutes $Al_1$; if negative, it supports $Al_1$.*

The knowledge in the example can be represented as an evidence structure as follows.

$\mathcal{L}_e = 2^{\{e_1, e_2\}};$

$\mathcal{L}_p = 2^{\{Al_1, Al_2, Al_3\}};$

$\mathcal{A} = \{\langle e_1, \{Al_1\}\rangle, \langle e_1, \{Al_2\}\rangle$
$\quad \langle \neg e_2, \{Al_1\}\rangle, \langle e_2, \{Al_2\}\rangle, \langle e_2, \{Al_3\}\rangle\};$

$\mathcal{R} = \phi \cup \mathcal{R}_\mathcal{A}$, where $\mathcal{R}_\mathcal{A}$ is the smallest relation on $\mathcal{A}$ satisfying the constraints for evidence structure. This means that we have no information about the strengths of the arguments (other than that derivable from the subset relations of $\mathcal{L}_e$ and $\mathcal{L}_p$).

Notice that in the representation above, evidential refutation "$e_2$, if positive, refutes $\{Al_1\}$" is represented as evidential supports "$e_2$, if positive, supports $\{Al_2\}$ as well as $\{Al_3\}$". The justification for doing so is that only *relative* evidential support strengths are important in $\mathcal{RES}$.

### 2.3 Evidence Accumulation

As noted before, all arguments are conditioned in the sense that argument $\langle e, p \rangle$ can enforce its support for $p$ only if $e$ is satisfied. This provides us with a natural way to accumulate evidence on an evidence structure.

Suppose our information allows us to grant a sentence $e$ in $\mathcal{L}_e$, then we can identify a set of arguments composed of all arguments whose presumptions are logical implications of $e$, i.e. the set of all arguments whose presumptions are granted if $e$ is granted and are thus applicable under $e$. This set and the relationships between arguments in this set, compose a sub-structure $\mathcal{ES}_e$ of $\mathcal{ES}$ for any sentence $e$ of $\mathcal{L}_e$. This sub-structure, defined as follows, is viewed as the result of evidence accumulation.

*Definition:* Let $\mathcal{ES} = \langle \mathcal{L}_e, \mathcal{L}_p, \mathcal{A}, \mathcal{R} \rangle$ be an evidence structure and $e$ be a sentence of $\mathcal{L}_e$. Then the sub-structure $\mathcal{ES}_e = \langle \mathcal{L}_{ee}, \mathcal{L}_{p_e}, \mathcal{A}_e, \mathcal{R}_e \rangle$, where

1. $\mathcal{L}_{ee} = \mathcal{L}_e$ ;
2. $\mathcal{L}_{p_e} = \mathcal{L}_p$ ;
3. $\mathcal{A}_e = \{\langle e', p \rangle | \langle e', p \rangle \in \mathcal{A}, e \rightarrow e'\}$;
4. $\mathcal{R}_e = \mathcal{R} \cap (\mathcal{A}_e \times \mathcal{A}_e)$, i.e. the sub-relation of $\mathcal{R}$ on $\mathcal{A}_e$,

is the *conditioned evidence structure* of $\mathcal{ES}$ conditioned on $e$.

The conditioning on an evidence structure has the following property.

**Corollary 1** *Let $\mathcal{ES} = \langle \mathcal{L}_e, \mathcal{L}_p, \mathcal{A}, \mathcal{R} \rangle$ be an evidence structure and $e_1, e_2 \in \mathcal{L}_e$ such that $e_1 \rightarrow e_2$. Then*

$$\mathcal{A}_{e_2} \subseteq \mathcal{A}_{e_1}.$$

**Proof**

The proposition is obvious because any argument triggered by $e_2$ will also be triggered by $e_1$. ∎

This proposition says that the set of triggered arguments will not contract with more evidence, i.e. the triggered argument set will expand monotonically with more evidence. But notice that $\mathcal{ES}_{e_1 \wedge e_2} = (\mathcal{ES}_{e_1})_{e_2}$ is not generally true, which means that evidence accumulation cannot be done cumulatively. Also notice that, as can be seen later, $\mathcal{RES}$ is non-monotonic of the conclusions it reached.

For our example presented in the last section, the evidence accumulation process is to conduct the tests and condition the evidence structure on the results of these tests. Different results will then issue different structures. The resulting conditioned evidence structures with different results of the tests are shown as follows. The $\mathcal{L}_e, \mathcal{L}_p$ and $\mathcal{R}$ parts are omitted because the former two parts are the same as in the original evidence structure, while the last part is empty except for these relationships derivable from subset relations.

$e_1$ **negative and** $e_2$ **negative**

$$\mathcal{A} = \{\langle \neg e_2, \{Al_2\}\rangle, \langle \neg e_2, \{Al_3\}\rangle\};$$

$e_1$ **negative and** $e_2$ **positive**

$$\mathcal{A} = \{\langle e_2, \{Al_1\}\rangle\};$$

$e_1$ **positive and** $e_2$ **negative**

$$\mathcal{A} = \{\langle e_1, \{Al_1\}\rangle, \langle e_1, \{Al_2\}\rangle,\\ \langle \neg e_2, \{Al_2\}\rangle, \langle \neg e_2, \{Al_3\}\rangle\};$$

$e_1$ **positive and** $e_2$ **positive**

$$\mathcal{A} = \{\langle e_1, \{Al_1\}\rangle, \langle e_1, \{Al_2\}\rangle, \langle e_2, \{Al_1\}\rangle\}.$$

### 2.4  Decision Making

Based on an evidence structure $\mathcal{ES}$ and evidence $e$, a partial relation on $\mathcal{L}_p$ can be defined with respect to how well a sentence is supported in $\mathcal{ES}_e$.

*Definition:* Let $\mathcal{ES} = \langle \mathcal{L}_e, \mathcal{L}_p, \mathcal{A}, \mathcal{R} \rangle$ be an evidence structure and $e$ an item of evidence, i.e. a sentence in $\mathcal{L}_e$. The comparison relation $\mathcal{C}$ on $\mathcal{L}_p$ determined by $\mathcal{ES}_e$, denoted by "$p_1 \leq p_2$", where $p_1$ and $p_2$ are two sentences in $\mathcal{L}_p$, is defined as:

1. in the case where there are arguments in $\mathcal{A}_e$ supporting some sentences $p'_1$ such that $p'_1 \rightarrow p_1$, then

$$p_1 \leq p_2 \stackrel{\text{def}}{=} \forall \langle e_1, p'_1 \rangle \in \mathcal{A}_e, ((p'_1 \rightarrow p_1) \rightarrow (\exists \langle e_2, p'_2 \rangle \in \mathcal{A}_e, (p'_2 \rightarrow p_2 \wedge \langle e_1, p'_1 \rangle \preceq \langle e_2, p'_2 \rangle)));$$

2. in the case where there are no arguments supporting any sentence $p'_1$ such that $p'_1 \rightarrow p_1$, then

$$p_1 \leq p_2 \stackrel{\text{def}}{=} \exists \langle e_2, p_2 \rangle \in \mathcal{A}_e, (p'_2 \rightarrow p_2.)$$

Informally, the first formula states that if there are arguments supporting $p_1$ or its subsets, i.e. there are justifications for conclusion $p_1$, then we can say $p_1$ is no more believable than $p_2$ if and only if for every justification for $p_1$, there exists at least one justification for $p_2$ and this justification (argument) is no worse than that justification for $p_1$. The second formula states that if there is no justification for conclusion $p_1$, then we can say $p_1$ is no more believable than $p_2$ if and only if $p_2$ has some justifications.

Using this relation, conclusions with respect to which conclusion is better supported than another is clarified. Intuitive terms can be defined.

*Definition:* Let $\mathcal{C}$ be a comparison relation on $\mathcal{L}_p$ and let $p_1, p_2$ be two sentences in $\mathcal{L}_p$, then we say

- $p_1$ *is less believable than* $p_2$ iff

$$p_1 \leq p_2 \text{ and } p_2 \not\leq p_1;$$

- $p_1$ *is as believable as* $p_2$ iff

$$p_1 \leq p_2 \text{ and } p_2 \leq p_1;$$

- $p_1$ *is not comparable to* $p_2$ iff

$$p_1 \nleq p_2 \text{ and } p_2 \nleq p_1.$$

From the relation so defined, decisions can be made. For example, we can have the following definition.

*Definition:* A conclusion $p$ is call *a plausible conclusion* supported by an evidence structure under evidence $e$ if and only if $\neg p$ is less believable than $p$.

Many other ways of decision making are also possible and sensible. For example, "choosing the best" from competitors (as we used in the examples in this paper), and/or providing some standard arguments as thresholds.

But more than that, why and how the decision is reached can be easily explicated in $\mathcal{RES}$ because of its symbolic representation. We can trace all support commanded by the competing choices to provide the reason why some of these supports are overwhelmed while others are taken as the basis for reaching the conclusion. This will provide justification for the results in cases where the results alone are not convincing enough.

Returning to our example above, for which we have built the evidence structure (section 2.2) and shown the results of evidence accumulation, *i.e.* conditioned evidence structures, with different results of the tests (section 2.3). From these conditioned evidence structures, relationships among conclusions with respect to evidential support under different forms of evidence (different results of the tests) are shown in the diagrams below (for clarity, the subset relation is shown only in the first case). In the diagrams, nodes are statements in the conclusion space and two nodes, one above another, have a path linking them iff the higher node is no less believable than the lower node.

$e_1$ **negative and** $e_2$ **negative** The diagram for this case is as follows, which shows that $\{Al_1\}$ is more believable than $\{Al_2, Al_3\}$ while the $\{Al_2\}$ and $\{Al_3\}$ are not comparable. This is so because the only argument applicable (with its presumption satisfied by the results of the tests) is $\langle \neg e_2, Al_1 \rangle$. This argument gives some support but only to $Al_1$.

Other relationships like $\{Al_1\}$ is more believable than $\{Al_2\}$ and $\{Al_3\}$ are also represented.

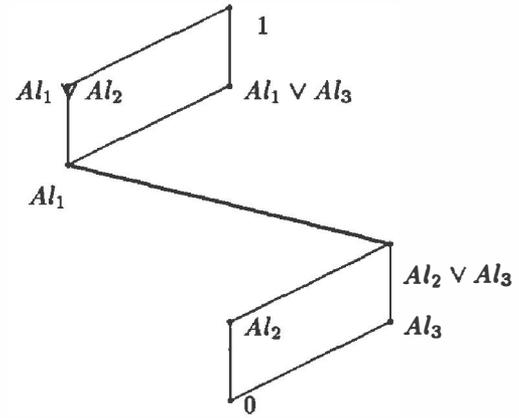

$e_1$ **negative and** $e_2$ **positive** In this case, $Al_1$ is refuted by $e_2$, which makes $\{Al_1\}$ less believable than either $\{Al_2\}$ or $\{Al_3\}$. So we have the following diagram (in which the subset relationships are omitted).

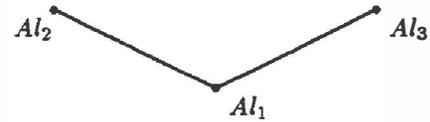

$e_1$ **positive and** $e_2$ **negative** In this case, $Al_1$ is supported by both $e_1$ and $\neg e_2$; $Al_2$ is supported only by $e_1$; and $Al_3$ commands no support at all. This gives us the following diagram.

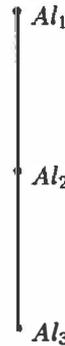

**Both $e_1$ and $e_2$ are positive** In this case, $Al_2$ is supported by $e_1$; although $Al_1$ is also supported by $e_1$, it is also refuted by $e_2$. This makes it less believable than $Al_2$ but not comparable with $Al_3$, which has no support.

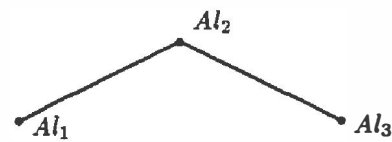

It can be seen that if we also know that $e_1$ is more trustworthy than $e_2$, (or *vice versa*), then in the last case we will able to say that $Al_1$ is more believable than $Al_3$ (or *vice versa*). It is something of a surprise that even this information of relative strength is not always needed, *e.g.* in the example above, this information makes no difference in the first three cases.

## 3  An Example

In this section, using $\mathcal{RES}$, we re-work the example of "The Hominids of East Turkana" which is the key example in both Shafer [10] and Cohen [4]. The exposition of the example from Shafer's paper is copied here.

### 3.1  The Example

**Example 2** In the August, 1978 issue of *Science America*, Alan Walker and Richard E.T. Leakey [11] discuss the hominids that have recently been discovered in the region east of Lake Turkana in Kenya. These fossils, between a million and two million years of age, show considerable variety, and Walker and Leakey are interested in deciding how many distinct species they represent.

In Walker and Leakey's judgment, the relatively complete cranium specimens discovered in the upper member of the Koobi Fora Formation in East Turkana are of three forms: (I) A "robust" form with large cheek teeth and massive jaws. These fossils show wide-fanning cheekbones, very large molar and premolar teeth, and smaller incisors and canines. The brain cases have an average capacity of about 500 cubic centimetres, and there is often a bony crest running fore and aft across its top, which presumably provided greater area for the attachment of the cheek muscles. Fossils of this form have also been found in South Africa and East Asia, and it is generally agreed that they should all be classified as members of the species *Australopithecus robustus*. (II) A Smaller and slenderer (more "fragile") form that lacks the wide-flaring cheekbones of I, but has similar cranial capacity and only slightly less massive molar and premolar teeth. (III) A large-brained (c. 850 cubic cm) and small-jawed form that can be confidently identified with the *Homo erectus* specimens found in Java and northern China.

The placement of the three forms in the geological strata in East Turkana shows that they were contemporaneous with each other. How many distinct species do they represent? Walker and Leakey admit five hypotheses:

1. I, II, III are all forms of a single, extremely variable species.

2. There are two distinct species: one *Australopithecus robustus*, has I as its male form and II as its female form; the other, *Homo erectus*, is represented by III.

3. There are two distinct species: one, *Australopithecus robustus*, is represent by I; the other has III, the so-called *Homo erectus* form, as its male form, and II as its female form.

4. There are two distinct species: one is represented by the fragile form II; the other, which is highly variable, consists of I and III.

5. The three forms represent three distinct species.

Here are the items of evidence, or arguments, that Walker and Leakey use in their qualitative assessment of the probabilities of these five hypotheses:

1. Hypothesis 1 is supported by general theoretical arguments to the effect that distinct hominid species cannot co-exist after one of them has acquired culture.

2. Hypothesis 1 and 4 are doubtful because they postulate extreme adaptations within the same species: The brain seems to overwhelm the chewing apparatus in III, while the opposite is true in I.

3. There are difficulties in accepting the degree of sexual dimorphism postulated by hypotheses 2 and 3. Sexual dimorphism exists among living anthropoids, and there is evidence from elsewhere that hints that dental dimorphism of the magnitude postulated by hypothesis 2 might have existed in extinct hominids. The dimorphism postulated by hypothesis 3, which involves females having roughly half the cranial capacity of males, is less plausible.

4. Hypotheses 1 and 4 are also impugned by the fact that specimens of type I have not been found in Java and China, where specimens of type III are abundant.

5. Hypotheses 2 and 3 are similarly impugned by the absence of specimens of type II in Java and China.

Before specimens of type III were found in the Koobi Fora Formation, Walker and Leakey thought it likely that I and II specimens constituted a single species. Now on the basis of the total evidence, they consider hypothesis 5 the most probable.

### 3.2  $\mathcal{RES}$ Design

Following the terminology of Shafer and Tversky, we will call our representation of the example using $\mathcal{RES}$ as a "design of evidence". The design is as follows.

- $\mathcal{L}_e = 2^{\{e_1, e_2, e_{12}, e_{23}, e_{13}\}}$, where

    $e_1$: "the theory";
    $e_2$: "Absence of type I and II among III in Far East";
    $e_{ij}$: "Difference between type $i$ and type $j$".

- $\mathcal{L}_p = \{B_1, B_2, B_3, B_4, B_5\}$, where

    $B_1$ = One species;
    $B_2$ = Two species, III and one composed of I (male) and II(female);
    $B_3$ = Two species, I and one composed of III (male) and II (female);
    $B_4$ = Two species, II and one composed of I and III;
    $B_5$ = Three species.

- Let
$$\mathcal{A}' = \left\{ \begin{array}{l} \langle e_1, B_1 \rangle, \langle e_2, B_2 \rangle, \langle e_2, B_5 \rangle, \\ \langle e_{12}, B_3 \rangle, \langle e_{12}, B_4 \rangle, \langle e_{12}, B_5 \rangle, \\ \langle e_{23}, B_2 \rangle, \langle e_{23}, B_4 \rangle, \langle e_{23}, B_5 \rangle, \\ \langle e_{13}, B_2 \rangle, \langle e_{13}, B_3 \rangle, \langle e_{13}, B_5 \rangle \end{array} \right\},$$
we have
$$\mathcal{A} = \mathcal{A}' \cup \{\langle e' \wedge e'', p \rangle | \langle e', p \rangle \in \mathcal{A}', \langle e'', p \rangle \in \mathcal{A}'\}.$$

Notice we convert arguments against some statements to be arguments supporting the complementary statements, as we did in the example in the previous sections, and we have not considered arguments with disjuncted presumptions.

Also consider $\mathcal{A} - \mathcal{A}'$. Adding this subset of arguments to $\mathcal{A}$ is equivalent to adding a rule "if both $e'$ and $e''$ individually support $p$, then $e' \wedge e''$ supports p". There are counter examples for this rule. Since we have not discussed how general this rule is and how well it can be justified, we did not put this rule as a requirement for $\mathcal{RES}$.

- $\mathcal{R} = \{e_{12} \prec e_1, e_1 \prec e_2 \wedge e_{13},$
  $\phantom{\mathcal{R} = \{} e_{12} \prec e_{13}, e_{23} \prec e_{13}\}$
  $\cup \{\langle e, p' \rangle = \langle e, p'' \rangle | \langle e, p' \rangle, \langle e, p'' \rangle \in \mathcal{A}\},$
where $e' \prec e''$ denotes
$\forall \langle e', p' \rangle, \langle e'', p'' \rangle \in \mathcal{A},$
$\phantom{\forall} ((\langle e', p' \rangle \preceq \langle e'', p'' \rangle) \wedge (\langle e'', p'' \rangle \npreceq \langle e', p' \rangle)).$

The reasons for these relationships in $\mathcal{R}$ can be found from the exposition of the example.

- The first relationship is based on the situation before III has been found; the theoretical support for $B_1$ is overwhelming. After III was found, this support is overwhelmed by new considerations (the second relationship). Obviously, the condition $e_2 \wedge e_{13}$ is sufficient to make the choice of conclusion as can be seen from the original example. But this condition might be not necessary. It might be true that some weaker conditions, e.g. $e_{23} \wedge e_2$; $e_{13}$ only; or even $e_{23}$ alone can also overwhelm $e_1$ and make the choice of conclusion. But where exactly the breaking point lies cannot be determined form the example and we are not in a position to judge this.
- The third and fourth relationships are based on the fact that the bigger the difference, the stronger this difference supports "different species".
- The last part of $\mathcal{R}$ is equivalent to adding a rule "if a piece of evidence supports many hypotheses, it supports them equally". This is not a very general case. For example, a piece of statistical evidence hardly supports any different hypotheses equally. In our example, we have many equivalences such like $\langle e_{12}, B_3 \rangle = \langle e_{12}, B_4 \rangle$ which is because $B_3$ and $B_4$ reflect the difference between I and II equally. It just so happens that we can put these relationships concisely.

Using the evidence structure built above, evidence can be accumulated and hypotheses can be chosen.

**Before III was found** It can be seen that the only arguments that can be triggered are those with presumption of $e_1$ or $e_{12}$. While $e_1$ is rendered a better evidence, the conclusion it supports, i.e. $B_1$, is the best supported hypothesis. Notice that since III has not been found yet, $B_1$ is equivalent to the statement "I and II form one species".

**After III was found** This time, all arguments are triggered. Considering those arguments supporting $B_2$, $B_3$, and $B_4$, we can find that for every such argument, there is an argument with equal weight supporting $B_5$, but not otherwise. Thus we know $B_5$ is more believable than $B_2$, $B_3$, and $B_4$. At the same time, there is an argument $\langle e_2 \wedge e_{13}, B_5 \rangle$ supporting $B_5$ which is strong enough to overwhelm the only argument $\langle e_1, B_1 \rangle$ supporting $B_1$. Thus, $B_5$ is also more believable than $B_1$ and so $B_5$ is the best choice.

Notice that we are also able to say "$B_2$ is more believable than $B_3$". This is true not only because $B_2$ commands support from $e_2$ while $B_3$ doesn't (as Cohen noticed) but also because $B_2$ commands support from $e_{13}$ which is stronger than the support $B_3$ commands from $e_{23}$ and $e_{12}$.

This concludes our presentation of the $\mathcal{RES}$ design for this example. In the rest of this section, we briefly compare the $\mathcal{RES}$ design with the designs of Shafer and Cohen.

### 3.3  Comparison to other methods

It can be noticed that representing strengths of arguments by relative strength is at an abstraction level between the level of addressing factors (reasons or endorsements) and the level of using numbers. This difference of levels leads us to the following points.

In Shafer's paper [10], two designs, a Bayesian design (conditioning design, in contrast with total design) and a belief function design are presented. The belief function design turns out to be more intuitive or even more correct in a sense. But it is obvious that the numbers, both the probabilities and belief function values in the designs, are given quite arbitrarily. One is tempted strongly to question their relevance as well as their correctness.

Apart from the arbitrary nature of the numbers, their high abstraction level also makes these two designs incapable of reflecting some subtle but interesting inferences. For example, as we noticed earlier, it might so happen that only a difference as big as that between I and III is strong enough to overrule our faith in the theory, so that $B_3$, honouring this difference, will be better than $B_1$. With the two designs of Shafer, we will have to change the numbers. But how can we change these numbers to reflect such subtle differences if these numbers are related to the situation we are considering *indirectly* and are assessed quite *arbitrarily*?

In Cohen's book [4], a work on evidential reasoning is reported. One explicitly stated purpose of his study is to see "how far one can go in evidential reasoning without measures of evidence" and the two basic observations on which his work is based are that "we know much about uncertainty besides belief measures" and "it is extremely difficult to assess these numbers." Trying to circumvent the problems, he avoids numbers by replacing them with "endorsements" which are records of reasons for believing or disbelieving a hypothesis.

The example above serves as a main example for his exposition of the endorsement model and is worked out as another design using the endorsement model where the numbers in Shafer's two designs are replaced by endorsements (reasons *pro* or *con* different hypotheses).

The obvious problem with this design is its complexity. The design is capable of reflecting some subtle reasoning, but it has the limitation that it has no facilities to specify the strengths of arguments. For this example, using the endorsement model we cannot capture Walker and Leakey's original statement that "The second hypothesis, we think, is more likely to be correct than the third". This, as we have already seen, has been reflected elegantly in $\mathcal{RES}$.

In fact, this defect is noticed by the author himself, which leads him to say that "... an inability and unwillingness to specify the relative weight of endorsements can limit the usefulness of the endorsement-based approach". This poses an interesting question of how well the endorsement model goes together with $\mathcal{RES}$ because though we believe that comparison relationships between arguments form a suitable interface for evidential reasoning, information at a more detailed level might be interesting and helpful. However, it seems to us that such a detailed level as that provided by endorsements is not usually required. This is the case in the example.

## 4  Summary

In this paper, we have described a method for evidential reasoning based on representing arguments and reflecting the relative strengths of arguments. Examples are presented showing its usage.

We have achieved some other results with $\mathcal{RES}$ as well [1]. For example, we have formalized the method to be a logic in which the constraints are reflected as some axioms; we have implemented it; we have shown that when making choices from competitive alternatives based on available evidence is our only concern, $\mathcal{RES}$ is capable of representing all information conveyed by any belief functions; we have established that $\mathcal{RES}$ is capable of reflecting the belief construction process advocated in [9]; we have shown that absolute measurement [6] such as probabilities [5], belief degrees [8], and certainty degrees [12] can also be represented using $\mathcal{RES}$ by introducing some "canonical examples" [10] as representatives of such numbers [2]; and we have shown that $\mathcal{RES}$ can also represent many patterns of common sense reasoning.

Together those results show that $\mathcal{RES}$ is a flexible and potentially powerful representation method, complementary to other alternative methods.

## Acknowledgements

Thanks to Prof. J. Pearl at UCLA for his helps. Thanks to the referees for pointing out a mistake in the draft paper.